\documentclass[final,3p,times,twocolumn]{elsarticle}
\usepackage{amssymb}
\usepackage{amsmath}
\usepackage{subcaption}
\journal{Nuclear Physics B}

\begin{document}

\begin{frontmatter}



\title{FireRescue: A UAV-Based Dataset and Enhanced YOLO Model for Object Detection in Fire Rescue Scenes}


\author[label1]{Qingyu Xu} %% Author name
\author[label1]{Runtong Zhang} %% Author name
\author[label1]{Zihuan Qiu} %% Author name
\author[label1]{Fanman Meng\corref{mycorrespondingauthor}} %% Author name
\cortext[mycorrespondingauthor]{Corresponding author}
\ead{fmmeng@uestc.edu.cn}

\affiliation{organization={School of Information and Communication Enginnering, University of Electronic Science and Technology of China},%Department and Organization
            addressline={No. 2006, Xiyuan Avenue}, 
            city={Chengdu},
            postcode={611731}, 
            state={Sichuan},
            country={China}}

%% Abstract
\begin{abstract}
Object detection in fire rescue scenarios is importance for command and decision-making in firefighting operations. However, existing research still suffers from two main limitations. First, current work predominantly focuses on environments such as mountainous or forest areas, while paying insufficient attention to urban rescue scenes, which are more frequent and structurally complex. Second, existing detection systems include a limited number of classes, such as flames and smoke, and lack a comprehensive system covering key targets crucial for command decisions, such as fire trucks and firefighters. To address the above issues, this paper first constructs a new dataset named "FireRescue" for rescue command, which covers multiple rescue scenarios, including urban, mountainous, forest, and water areas, and contains eight key categories such as fire trucks and firefighters, with a total of 15,980 images and 32,000 bounding boxes. Secondly, to tackle the problems of inter-class confusion and missed detection of small targets caused by chaotic scenes, diverse targets, and long-distance shooting, this paper proposes an improved model named FRS-YOLO. On the one hand, the model introduces a plug-and-play multidi-mensional collaborative enhancement attention module, which enhances the discriminative representation of easily confused categories (e.g., fire trucks vs. ordinary trucks) through cross-dimensional feature interaction. On the other hand, it integrates a dynamic feature sampler to strengthen high-response foreground features, thereby mitigating the effects of smoke occlusion and background interference. Experimental results demonstrate that object detection in fire rescue scenarios is highly challenging, and the proposed method effectively improves the detection performance of YOLO series models in this context.
\end{abstract}


% \begin{highlights}
% \end{highlights}

%% Keywords
\begin{keyword}

dataset, object detection, attention, UAV, fire rescue

\end{keyword}

\end{frontmatter}



% \section{Example Section}
% \label{sec1}
%% Labels are used to cross-reference an item using \ref command.

\section{Introduction}
\label{sec1}

Against the backdrop of increasing frequency of global natural disasters, the importance of disaster rescue operations has become increasingly prominent\cite{xiong2023research}. As a critical component of the disaster rescue command system, UAV(Unmanned Aerial Vehicle) reconnaissance currently relies primarily on manual visual identification. This method is inefficient and struggles to provide commanders with comprehensive and accurate key command elements from the disaster scene, highlighting an urgent need for the development of efficient automatic identification technology. In this context, the development of highly accurate and easily deployable target detection methods for fire rescue scenarios has become a core task in enhancing the effectiveness of UAV reconnaissance. 

Nevertheless, the advancement of research in this field is severely constrained by a critical factor: the absence of high-quality, domain-specific benchmark datasets. Prevailing public drone datasets like VisDrone\cite{du2019visdrone} and SDD\cite{robicquet2016learning} exhibit significant limitations when applied to fire rescue scenarios, which can be summarized in two aspects: (1)Scenario Bias: Existing fire-related datasets captured by drones predominantly focus on forest fire scenarios, failing to adequately encompass the more frequent and hazardous building fire incidents. Moreover, they lack the visual complexity characteristic of real urban fire grounds, which are defined by dense smoke, erratic illumination, and cluttered backgrounds; (2) Category Mismatch: They lack essential object categories such as "fire truck", "firefighters",  "flames" and "smoke". Consequently, models trained on these generic datasets often demonstrate suboptimal and unstable performance when deployed in authentic, complex fire rescue environments.

First, from high-altitude UAV perspectives, critical targets such as fire trucks occupy minimal pixel areas within complex scenes, making traditional detection methods prone to misses. Second, intense visual interference—caused by dense smoke, flames, water mist, dust, and structural debris—results in blurred images, reduced contrast, and obscured features, imposing stringent demands on model robustness under extreme conditions. Third, rescue vehicles and ordinary trucks exhibit high similarity in color and shape; when densely parked, mutual occlusion often triggers cascading false detections, thereby challenging the model’s capability to distinguish similar objects in crowded layouts. These challenges necessitate detection models with enhanced feature extraction and multi-scale perception capabilities, for which current general purpose object detection frameworks remain inadequate.

To address the dual challenges in data and algorithms outlined above, this study undertakes a comprehensive research initiative. First, we construct "FireRescue", a large-scale, meticulously annotated dataset of aerial images from firefighting drones, designed to establish a reliable benchmark for the field. Furthermore, we propose a novel lightweight object detection framework, termed FRS-YOLO, which incorporates two key architectural improvements specifically tailored to overcome the technical difficulties identified.

The main contributions of this work are summarized as follows:

1. We pioneer the construction and open-source release of the FireRescue dataset, comprising 15980 images and 32000 instance annotations. It encompasses critical categories such as "Emergency Rescue Fire Truck","Water Tanker Fire Truck","Firefighter","Flames","Smoke" and accurately captures the complexity of real-world fire rescue scenarios, thereby advancing research in this domain.

2. We design a novel Multi-Dimensional Collaborative Enhancement Attention (MCEA) module. By parallelly integrating global max pooling(GMP), standard deviation pooling(GSP), average pooling(GAP), this module mitigates the insufficient feature response of traditional attention mechanisms in complex backgrounds, significantly boosting the model's capacity to capture discriminative features.

3. We integrate the MCEA module into the backbone network of YOLOv12, forming a new A2C2f-MCEA core component. Concurrently, we replace conventional upsampling with the dynamic sampling method "Dysample" to enhance the spatial reconstruction fidelity of feature maps, thereby improving the localization and identification of small and obscured objects.

4. We propose the complete FRS-YOLO model incorporating the aforementioned innovations and conduct extensive experiments on our self-developed FireRescue dataset. The results conclusively demonstrate that our approach significantly outperforms existing state-of-the-art methods in key metrics such as detection accuracy and recall, while maintaining model lightweightness.

The rest of this paper is organized as follows: Section 2 (Related Work), Section 3 (The Proposed FRS-YOLO Method), Section 4 (Experimental Results and Discussion), and Section 5 (Conclusion and Future Work).

%-------------------------------------------------------------------------
\section{Related Work}
\label{sec2}
\subsection{Specialized Object Detection Datasets}
High-quality object detection datasets serve as the cornerstone for algorithmic advancement in computer vision. While general-purpose datasets such as MS COCO\cite{lin2014microsoft} and Pascal VOC\cite{everingham2010pascal} have greatly propelled the development of object detection models, their limitations in terms of scene coverage and category design become evident when applied to vertical domains such as low-altitude UAV vision. This has motivated the creation of specialized datasets tailored to specific application contexts.

In the field of low-altitude vision, dedicated datasets are often characterized by strong scenario-specific and task-oriented attributes. For example, VisDrone\cite{du2019visdrone} has significantly promoted research on vehicle and pedestrian detection in UAV-captured imagery by providing a large-scale, multi-scenario benchmark. To address more specialized tasks, several targeted datasets have been introduced: AnimalDrone\cite{zhu2021graph} focuses on wildlife counting and monitoring in complex natural settings; DroneSwarms\cite{cao2024visible} addresses the challenge of small object detection in counter-UAV missions, with approximately 99.6\% of its objects smaller than 32 pixels; and VDD\cite{xiao2025uav} is designed specifically for vehicle detection and classification in aerial videos. These datasets demonstrate that precise scenario definition and granular annotation can substantially improve model performance within their respective domains.

However, as noted by \cite{Sun2025ASurveyforUAVs} in their recent survey, despite considerable progress in specialized dataset development, the critical public safety area of fire rescue still lacks a high-quality, publicly available benchmark. Existing datasets generally do not align with key fireground categories—such as firefighters, specialized fire apparatus, flames, and smoke—nor do they adequately represent challenging conditions like heavy smoke, open flames, and complex obstructions typical in real fire environments. This data gap has considerably hindered the application of computer vision to crucial fire rescue tasks including early fire warning, rescue command, and personnel positioning.

To address this shortage, this paper introduces FireRescue, a dedicated dataset designed for fire rescue scenarios. By integrating authentic fire operation footage with selected public data, it covers essential categories including multiple types of fire trucks, firefighters, flames, and smoke. The dataset is intended to serve as a reliable benchmark to foster research in visual perception for emergency response, facilitating the adoption of low-altitude intelligence technology in life-saving applications.

\subsection{General Object Detection Algorithms}

In recent years, computer vision has made remarkable progress in object detection, particularly with deep learning-based algorithms\cite{zhang2023few}, offering new approaches for fire rescue scene analysis. Current deep learning-based object detection methods are primarily categorized into one-stage and two-stage algorithms. Representative two-stage algorithms such as R-CNN\cite{yasir2024shipgeonet}, Fast R-CNN\cite{girshick2015fast}, and Faster R-CNN\cite{ren2016faster} rely on pre-generated region proposals for classification and localization. This design requires substantial computational resources and results in slower detection speeds, making them unsuitable for deployment in fire rescue scenarios where computational resources are constrained and real-time performance is critical. In contrast, typical one-stage algorithms like the YOLO series and SSD\cite{li2024ws} directly predict both object categories and bounding boxes from images. YOLOv1 to YOLOv3\cite{redmon2016you}\cite{redmon2017yolo9000} \cite{redmon2018yolov3} are the pioneers of YOLO-series models, and their performance improvements are all closely related to the backbone improvement and make DarkNet widely used. YOLOv4\cite{bochkovskiy2020yolov4} introduces a large number of residual structure design proposed CSPDarknet53 backbone network, which effectively reduces computational redundancy and realizes high-performance feature expression and efficient training. YOLOv5\cite{jocher2020ultralytics} employs a Focus module at the input stage that performs image slicing and splicing to preserve more feature information than direct convolutional downsampling. Additionally, it utilizes K-Means clustering to automatically generate anchor boxes prior to training, thereby reducing the need for manual parameter tuning. YOLOv6\cite{li2023yolov6} began to incorporate reparameterization. YOLOv7\cite{wang2023yolov7} proposes the E-ELAN structure to enhance the model capability without destroying the original gradient flow. YOLOv8\cite{jocherultralytics} combines the features of the previous generations of YOLOs and adopts the C2f structure with richer gradient streams, which is lightweight and adaptable to different scenarios while taking accuracy into account. The recent evolution of the YOLO series, particularly with the rise of deep learning and Transformer architectures, has witnessed integration attempts between self-attention mechanisms and convolutional neural networks (CNNs). Models such as YOLOv9\cite{wang2024yolov9}, YOLOv10\cite{wang2024yolov10}, YOLOv11\cite{khanam2024yolov11} and YOLOv12\cite{tian2025yolov12} reflect this trend by incorporating self-attention into their frameworks or adopting Transformer-based blocks in feature extraction, aiming to enhance representational power. YOLOv9 incorporates attention modules, such as the Convolutional Block Attention Module (CBAM) or Efficient Channel Attention (ECA), at critical positions within its backbone network, thereby enhancing the model's focus on target regions.YOLOv10 incorporates lightweight Transformer modules during the feature extraction stage, which can be fused with CNN components in either parallel or serial configurations. Additionally, it employs a more refined variant of the IoU loss function, such as EIoU, for bounding box regression. YOLOv11 incorporates a visualization module within its architecture to output attention heatmaps in real-time, and adapts anchor sizes and ratios dynamically during training to better accommodate the target distribution of different datasets. YOLOv12 deeply integrates multiple attention mechanisms, including self-attention, CBAM, and ECA, into both its backbone network and feature fusion layers. This multi-branch design enhances feature representation capabilities without significantly increasing computational complexity. Furthermore, the DETR series\cite{carion2020end}\cite{zhu2020deformable}\cite{zhang2022dino}\cite{zhao2024detrs} has pioneered the integration of Transformer architectures into object detection. However, these models still face challenges such as slow training convergence, high computational cost, and limited efficacy in small object detection. Consequently, the YOLO series maintains a leading position in the small-model domain due to its superior balance between accuracy and inference speed. While the aforementioned general-purpose detectors have achieved notable success on public benchmarks, their designs are inherently generic and often fall short of optimal performance when confronted with the unique challenges of specific vertical domains, such as fire rescue. This limitation motivates our exploration of specialized model designs tailored to domain-specific requirements.

\subsection{Application of Attention Mechanisms in Object Detection}
Attention mechanisms have been generalized to CNNs in the way of refining feature activations and have shown great potential in image recognition. For the first time, the most representative work, SE\cite{hu2018squeeze}, presents an effective method for learning channel attention while achieving notable performance. It first aggregates the spatial information with the help of 2D global average pooling and then utilizes two fully-connected layers with dimensionality reduction to capture inter-channel interactions. Inheriting the settings of Squeeze and Excitation in SE, later methods either put some effort into boosting the Squeeze phase (e.g., GSoP-Net\cite{gao2019global} and FcaNet\cite{qin2021fcanet}) or reduce the complexity of the Excitation phase by adopting a 1D convolution filter (e.g., ECA\cite{wang2023banet}), or attempt to improve all phases at the same time (e.g., SRM\cite{lee2019srm} and GCT\cite{yang2020gated}). Besides channel attention, spatial attention, regarded as an adaptive spatial region selection mechanism, plays another vital role in inferring fine attention. CBAM\cite{woo2018cbam} and BAM\cite{park2018bam} provide robust representative attention by effectively exploiting the advantages of channel and spatial attention. However, mechanisms such as Squeeze-and-Excitation (SE) and Convolutional Block Attention Module (CBAM) suffer from non-trivial computational overhead and the potential loss of deep visual representations caused by channel dimensionality reduction. The Multi-dimensional Collaborative Attention module (MCAM)\cite{yu2023mca} was proposed to address these limitations. Its lightweight design facilitates synergistic modeling across the channel, height, and width dimensions, enabling the dynamic capture of critical features while maintaining high computational efficiency. While existing attention mechanisms like CBAM and MCAM have explored the combination of various pooling operations, they primarily employ Global Average Pooling (GAP), Standard Deviation Pooling(GSP), Global Max Pooling (GMP) in separate attention sub-networks (i.e., channel and spatial). This design may dilute the strength of critical features against complex backgrounds. In contrast, our proposed Multi-dimensional Collaborative Enhancement Attention (MCEA) module innovatively integrates GAP, GSP, GMP through parallel fusion during the "squeeze" phase of channel attention. This integration aims to generate a more comprehensive and discriminative channel descriptor, thereby enhancing the model's feature selection capability in cluttered scenes more directly and effectively.

\section{Methodology}
\label{sec:formatting}

This chapter is structured as follows: First, it justifies the need for a custom dataset and outlines the construction pipeline of FireRescue; Second, it provides a high-level overview of the FRS-YOLO architecture; Finally, it offers a detailed analysis of the design and implementation of two key improvement modules.

%-------------------------------------------------------------------------
\subsection{The FireRescue Dataset}
\subsubsection{Construction Motivation and Domain Gap}
As highlighted by Sun et al.\cite{Sun2025ASurveyforUAVs} in a related survey, although low-altitude vision datasets have developed rapidly, dedicated benchmark datasets for the critical vertical domain of fire rescue remain notably scarce. Existing general datasets (e.g., VisDrone, UAVDT), while covering some urban scenarios, generally lack key categories such as "Firefighter," "Specialized Fire Truck," "Flames," and "Smoke." Moreover, they struggle to realistically simulate dynamic disturbances common in fireground environments, such as dense smoke, open flames, and complex occlusions. To address this, this paper constructs the FireRescue dataset, aiming to precisely fill this domain gap and provide essential data support for the development and evaluation of intelligent perception algorithms in fire rescue scenarios.
\subsubsection{Data Sources and Collection Challenges}
The FireRescue dataset integrates multiple data sources. Its primary component originates from real fireground and training exercise footage provided by fire rescue departments, supplemented by carefully selected public fire and smoke datasets for effective expansion. It should be noted that as some imagery is sourced from historical internet data or various UAV platforms, the specific camera models and shooting parameters are difficult to trace uniformly. This objective circumstance inherently enhances the dataset's heterogeneity, posing greater demands on the model's generalization capabilities across scenes and devices. The data content encompasses significant illumination variations, substantial target scale changes, and complex background interference from the UAV perspective, thereby providing rich learning samples to address visual challenges in real rescue environments.
\subsubsection{Data Annotation and Quality Assurance}
The dataset comprises a total of 15,980 images and over 32,000 instance annotations, covering 8 categories of typical key targets at fire rescue scenes. These include 5 types of common specialized fire trucks, firefighters, flames, and smoke, among others. All images were meticulously annotated using the professional tool "Anylabeling". 

To ensure the annotations possess both visual accuracy and domain expertise, we implemented a rigorous quality control process: all annotation results underwent frame-by-frame review and correction by a dedicated team of 9 senior firefighting experts. All team members possess more than 5 years of frontline firefighting and rescue experience and hold the National Level 1 Certified Fire Engineer. Some members also have backgrounds in computer science. This "human-machine collaboration, expert verification" quality control mechanism fundamentally guarantees the semantic correctness and reliability of the annotations in complex scenarios.
\subsubsection{Dataset Statistics and Splits}
\begin{figure}[ht]
  \centering
\includegraphics[width=0.8\linewidth]{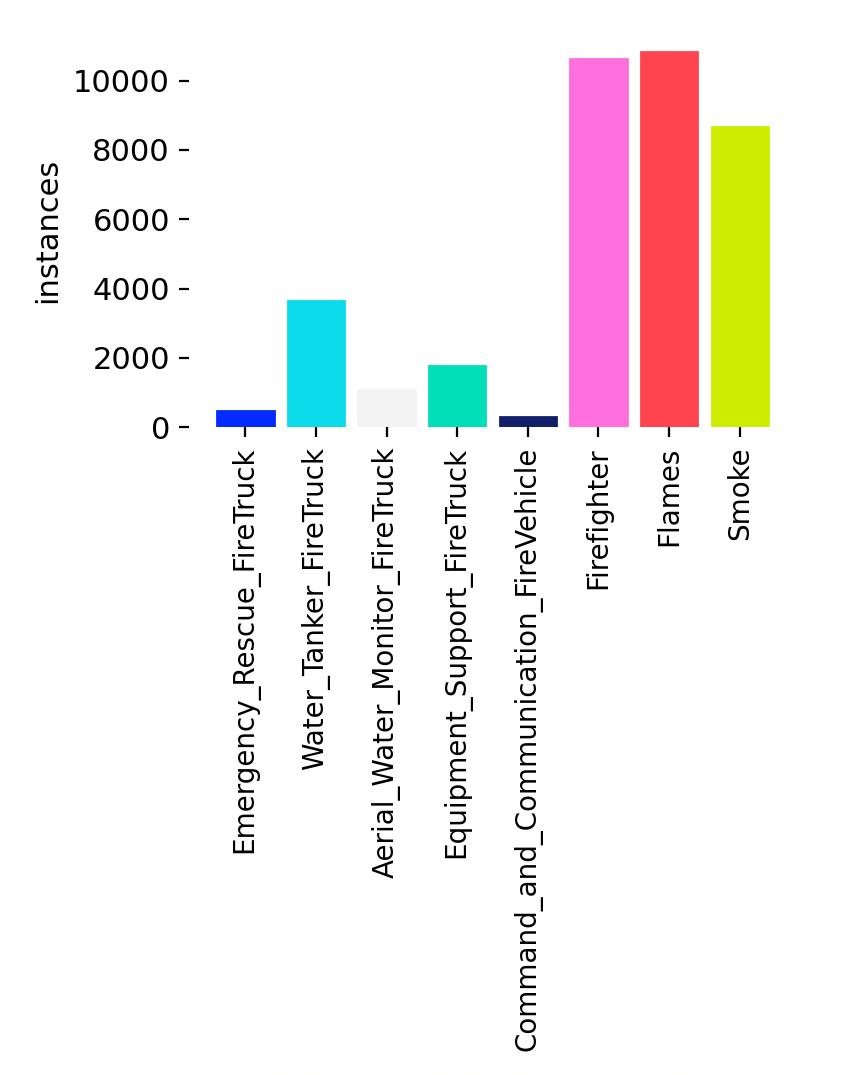}
  \caption{Annotation Statistics of the FireRescue Dataset}
  \label{fig:label-statistics}
\end{figure}

The detailed categories and distribution of instance counts within the dataset are shown in Fig. \ref{fig:label-statistics}. The scenes cover urban roads, suburban environments, and night scenes under various weather conditions, including numerous small and occluded targets, making the dataset suitable for research tasks such as small object detection and robustness testing in complex scenes. Finally, we randomly split the dataset into training sets, validation sets and test sets in a 8:1:1 ratio to ensure the fairness and reproducibility of experimental evaluations.
\vspace{5pt}

\subsection{The FRS-YOLO Architecture}

\begin{figure*}[ht]
  \centering
\includegraphics[width=\linewidth]{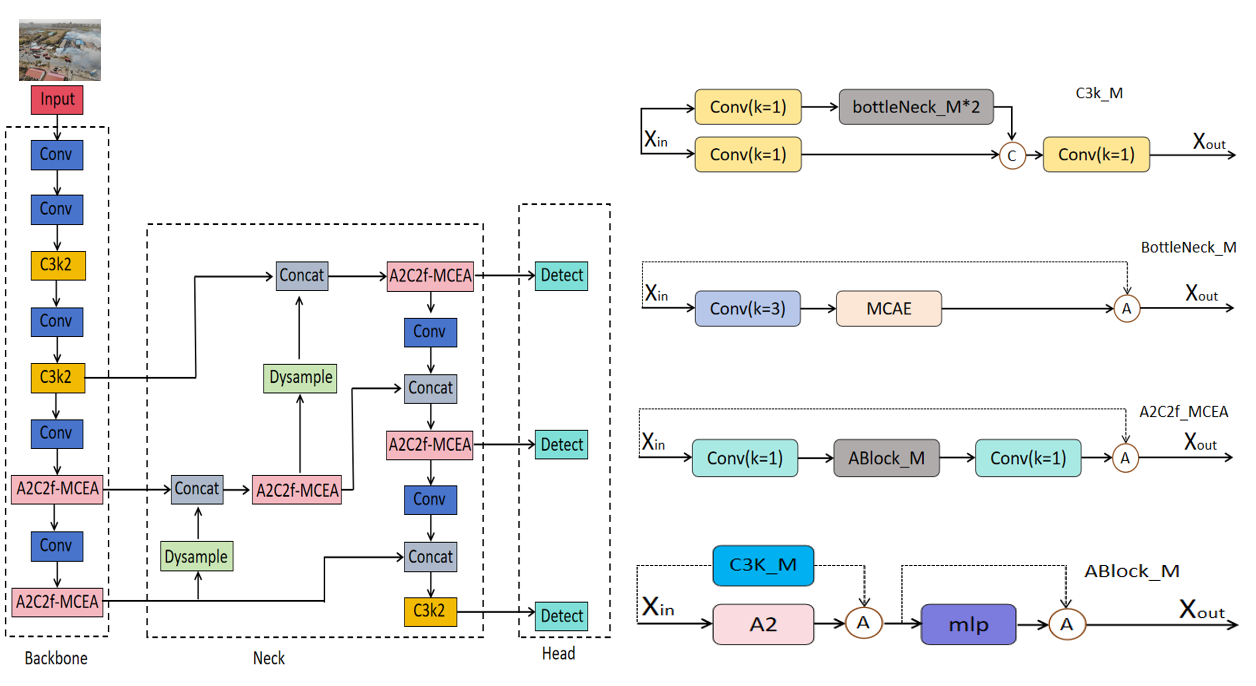}
  \caption{Structure of FRS-YOLO}
  \label{fig:FRS-YOLO-structure}
\end{figure*}
\vspace{5pt}

Leveraging the visual characteristics presented by the FireRescue dataset, this study aims to design a detector that pursues dual objectives: high accuracy and lightweight efficiency. This section details the overall architecture of FRS-YOLO. As illustrated in Figure \ref{fig:FRS-YOLO-structure}, FRS-YOLO retains the fundamental Backbone-Neck-Head structure of the baseline YOLOv12 model. Our enhancements are primarily two-fold: First, the integration of our proposed Multi-Dimensional Collaborative Enhancement Attention (MCEA) module into the backbone network, forming an A2C2f-MCEA block(Figure\ref{fig:FRS-YOLO-structure}); Second, the replacement of the conventional upsampling operation in the neck network with the dynamic upsampler, Dysample.

\subsection{Multi-Dimensional Collaborative Enhancement Attention Module}
\subsubsection{The Architecture of the MCEA Module}

\begin{figure*}[ht]
  \centering
  \includegraphics[width=0.9\linewidth, height=0.8\textheight, keepaspectratio]{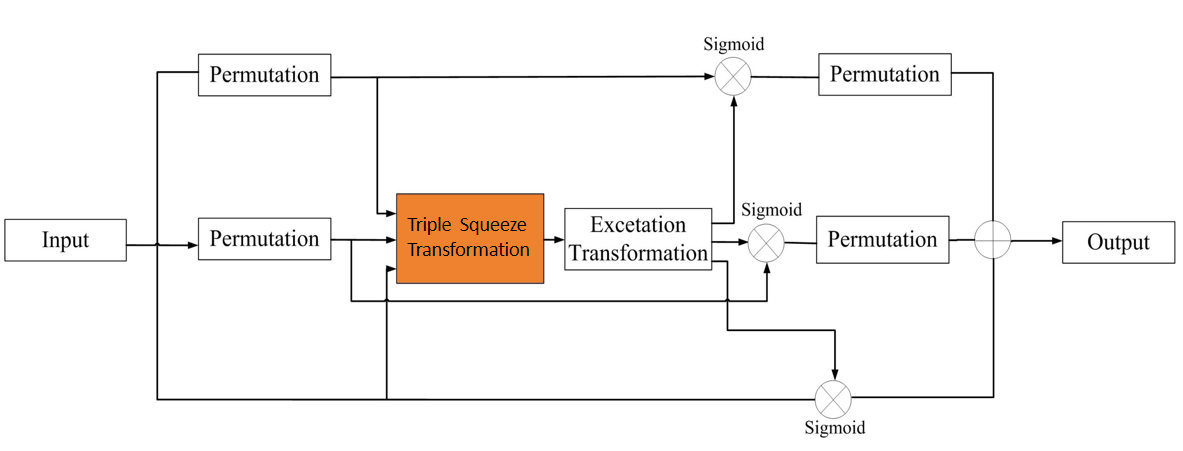}
  \caption{Structure of MCEA}
  \label{fig:MCEA}
\end{figure*}
\vspace{5pt}

We propose a Multi-dimensional Collaborative Enhancement Attention (MCEA) module, which builds upon and improves the classical MCA structure [27]. The overall framework, illustrated in Figure \ref{fig:MCEA}, consists of three parallel branches: width, height, and channel. The core squeeze transformation within each branch will be elaborated in the following subsection.

The MCEA module consists of three parallel branches(width-wise branch, height-wise branch, channel-wise branch). Let $\mathbf{F} \in \mathbb{R}^{C \times H \times W}$ be the output of the convolutional layer and subsequently the feature map fed into the MCA module, where $C$, $H$ and $W$ denote the number of channels (i.e., the number of filters), height, and width of the spatial feature map, respectively. The input feature map $\mathbf{F}$  is first processed through three parallel branches: H-axis rotation $PM_{H}()$, W-axis rotation $PM_{W}()$, and identity transformation $IM$, yielding feature maps $\widehat{\mathbf{F}}_{W} \in \mathbb{R}^{W \times H \times C}$, $\widehat{\mathbf{F}}_{H} \in \mathbb{R}^{H \times C \times W}$, $\widehat{\mathbf{F}}_{C} \in \mathbb{R}^{C \times H \times W}$, respectively. Each of these feature maps then undergoes a triple squeeze transformation $T_{tsq}()$ to produce $\hat{\mathbf{F}}_{W} \in \mathbb{R}^{W \times 1 \times 1}$, $\hat{\mathbf{F}}_{H} \in \mathbb{R}^{H \times 1 \times 1}$, $\hat{\mathbf{F}}_{C} \in \mathbb{R}^{C \times 1 \times 1}$. These are subsequently processed through an excitation transformation $T_{tsq}()$, an activation function $\sigma()$ and a rotation operation $PM^{-1}()$, ultimately generating feature maps $\mathbf{F}^{\prime\prime}_{W}$, $\mathbf{F}^{\prime\prime}_{H}$, $\mathbf{F}^{\prime\prime}_{C}$ that maintain the same dimensions as the original input. Mathematically, this process for the width branch $\mathbf{F}^{\prime\prime}_{W}$ can be summarized as the following equations. Similarly, the operations for $\mathbf{F}^{\prime\prime}_{H}$ and  $\mathbf{F}^{\prime\prime}_{C}$ follow analogous formulations.
\begin{equation}
\widehat{\mathbf{F}}_{W}=PM_{H}(\mathbf{F})
\label{eq:H-rotation}
\end{equation}

\begin{equation}
\hat{\mathbf{F}}_{W}=T_{tsq}(\widehat{\mathbf{F}}_{W}),
\tilde{\mathbf{F}}_{W}=T_{ex}(\hat{\mathbf{F}}_{W})
\label{eq:H-rotation}
\end{equation}

\begin{equation}
\mathcal{A}_{W}=\sigma(\tilde{\mathbf{F}}_{W}),
\tilde{\mathbf{F}}^{\prime}_{W}=\mathcal{A}_{W}\otimes\widehat{\mathbf{F}}_{W},
\tilde{\mathbf{F}}^{\prime\prime}_{W}=PM^{-1}(\tilde{\mathbf{F}}^{\prime}_{W})
\label{eq:H-rotation}
\end{equation}

Finally, the output feature maps from all three branches are fused through an aggregation operation to produce the final feature representation $\mathbf{F}^{\prime\prime}$.
\begin{equation}
\mathbf{F}^{\prime\prime}=\frac{1}{3}\otimes(\mathbf{F}_W^{\prime\prime}\oplus\mathbf{F}_H^{\prime\prime}\oplus\mathbf{F}_C^{\prime\prime})
\label{eq:fusion}
\end{equation}

\subsubsection{Triple Squeeze Transformation}

\begin{figure*}[ht]
  \centering
  \includegraphics[width=0.9\linewidth, height=0.8\textheight, keepaspectratio]{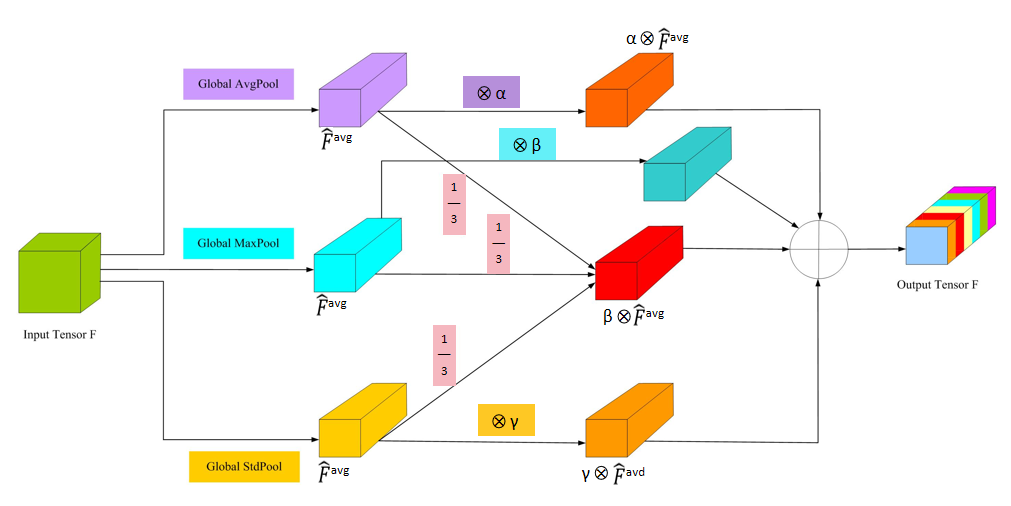}
  \caption{Structure of Triple Squeeze Transformation}
  \label{fig:tsq}
\end{figure*}
\vspace{5pt}

Although original MCA module module has achieved remarkable results through cross-dimensional interaction, we observe that its core component "Squeeze Transformation" to account for the specific characteristics of fire rescue scenarios, leaving untapped potential in its aggregation of global spatial information. Specifically, the underutilization of global pooling operations limits its capacity to capture complex spatial structural distributions. Inspired by the MCAM's approach of using average pooling and standard deviation pooling to aggregate cross-dimensional feature responses, we propose further enhancements tailored to the imaging properties of fire rescue elements (such as fire trucks and firefighters) from a UAV perspective—namely, their small pixel footprint, complex backgrounds, and susceptibility to occlusion. We argue that while average pooling captures global context and standard deviation pooling reflects internal feature variations, both exhibit limitations in identifying decisive local features of small targets. Based on this, we have designed a “Triple Squeeze Transformation” $T_{tsq}()$ (Figure \ref{fig:tsq}) component within the MCEA module.  By introducing a global max pooling branch, it directly identifies sparse yet highly responsive key pixels (such as the red color of a fire truck or the outline of a warning light), effectively preventing the features of small targets from being "diluted" by complex backgrounds. This provides the model with the most prominent discriminative evidence, significantly enhancing the robustness and accuracy of small target detection.

Given the input feature map $\mathbf{\widehat{F}}_{W}=\left[\mathbf{\hat{f}}_{1},\mathbf{\hat{f}}_{2},\ldots,\hat{\mathbf{f}}_{W}\right]\in\mathbb{R}^{W\times H\times C}$, we first employ global average pooling, standard deviation pooling, global maximum pooling to aggregate spatial information, generating three distinct channel-wise feature statistics, namely $\hat{\mathbf{F}}_{W}^{avg}=\left[\hat{f}_{1}^{\mathrm{avg}},\hat{f}_{2}^{avg},\ldots,\hat{f}_{W}^{\mathrm{avg}}\right]\in\mathbb{R}^{W\times1\times1}$ , $\hat{\mathbf{F}}_{W}^{\mathrm{std}}=\left[\hat{f}_{1}^{\mathrm{std}},\hat{f}_{2}^{std},\ldots,\hat{f}_{W}^{\mathrm{std}}\right]\in\mathbb{R}^{W\times1\times1}$ and $\hat{\mathbf{F}}_{W}^{\mathrm{max}}=\left[\hat{f}_{1}^{\mathrm{max}},\hat{f}_{2}^{max},\ldots,\hat{f}_{W}^{\mathrm{max}}\right]\in\mathbb{R}^{W\times1\times1}$. They represent the average pooling feature descriptor, standard deviation pooling feature descriptor, global max pooling feature descriptor, respectively. Specifically, the three pooling operations for the m-th channel can be expressed as follows:
\begin{equation}
\hat{f}_m^{avg}=\frac{1}{H\times C}\sum_{i=1}^{H}\sum_{j=1}^{C}\hat{\mathbf{f}}_m(i,j)
\end{equation}
\begin{equation}
\hat{f}_m^{std}=\sqrt{\frac{1}{H\times C}\sum_{i=1}^{H}\sum_{j=1}^{C}\left(\mathbf{f}_m(i,j)-\hat{f}_m^{avg}\right)^2}
\end{equation}
\begin{equation}
\hat{f}_m^{max}=MAX\{\hat{\mathbf{f}}_m(i,j)\}
\end{equation}
Here, $\hat{\mathbf{f}}_m\in\mathbb{R}^{1\times H\times C}$ denotes the feature map of the m-th channel of the input $\mathbf{\hat{F}}_{W}$. $\hat{f}_m^{avg}$, $\hat{f}_m^{std}$, $\hat{f}_m^{max}$ represent the different output feature descriptors corresponding to the m-th channel, respectively. Subsequently, $\hat{\mathbf{F}}_{W}^{avg}$, $\hat{\mathbf{F}}_{W}^{std}$, $\hat{\mathbf{F}}_{W}^{max}$ are fed into our designed adaptive combination mechanism to generate the width-wise feature descriptor $\mathbf{\hat{F}}_{W}$. Mathematically, this process can be formulated as follows:

\begin{equation}
\begin{split}
\hat{\mathbf{F}}_{W} = &T_{tsq}\left(\hat{\mathbf{F}}_{W}\right)\\ 
= &\frac{1}{3} \otimes \left( \hat{\mathbf{F}}_{W}^{avg} \oplus \hat{\mathbf{F}}_{W}^{std} \oplus \hat{\mathbf{F}}_{W}^{max}\right) \\ 
&+ \alpha \otimes \hat{\mathbf{F}}_{W}^{avg} \oplus \beta \otimes \hat{\mathbf{F}}_{W}^{std} \oplus \gamma \otimes \hat{\mathbf{F}}_{W}^{max}
\end{split}
\end{equation}

Here, $T_{tsq}()$ represents the triple squeeze operation, while $\alpha$, $\beta$, $\gamma$ are three trainable floating-point parameters, all satisfying $0 < \alpha, \beta, \gamma < 1$ and can be optimized through Stochastic Gradient Descent (SGD). This adaptive mechanism introduces input-dependent dynamics, enabling the assignment of varying weights to average pooling features, standard deviation pooling features, global max pooling features at different stages of image feature extraction. Thereby, it enhances the discriminative power of the output feature descriptors. Similarly, based on the aforementioned discussion, we can derive the height feature descriptor $\mathbf{\hat{F}}_{H}$ and the channel feature descriptor $\mathbf{\hat{F}}_{C}$ in another two branches. The effectiveness of the proposed method in adaptively aggregating dual information is empirically confirmed, as detailed in Section 4.3.

\subsection{Dysample}

In the neck of the YOLOv12 architecture, the most commonly used upsampling methods are nearest-neighbor (NN) and bilinear interpolation. However, both approaches rely on fixed interpolation rules and lack adaptability to image content, which to some extent limits the model's feature reconstruction capability in complex scenarios. To address this limitation, this study introduces an innovative dynamic upsampler—Dysample\cite{liu2023learning}. This method abandons the complex kernel generation mechanism of traditional dynamic convolutions and adopts an efficient dynamic point sampling strategy, significantly simplifying the computational path and effectively reducing inference latency. Furthermore, while maintaining a lightweight structure, Dysample has been validated to achieve excellent performance across multiple dense prediction tasks. Its dynamic upsampling mechanism enables more accurate multi-scale feature fusion, which helps enhance the recognition and localization of critical targets—such as trapped individuals and hazardous sources—in complex fire rescue environments with smoke and debris interference, thereby effectively suppressing false positives and missed detections.

\begin{figure}[ht]
  \centering
\includegraphics[width=\linewidth]{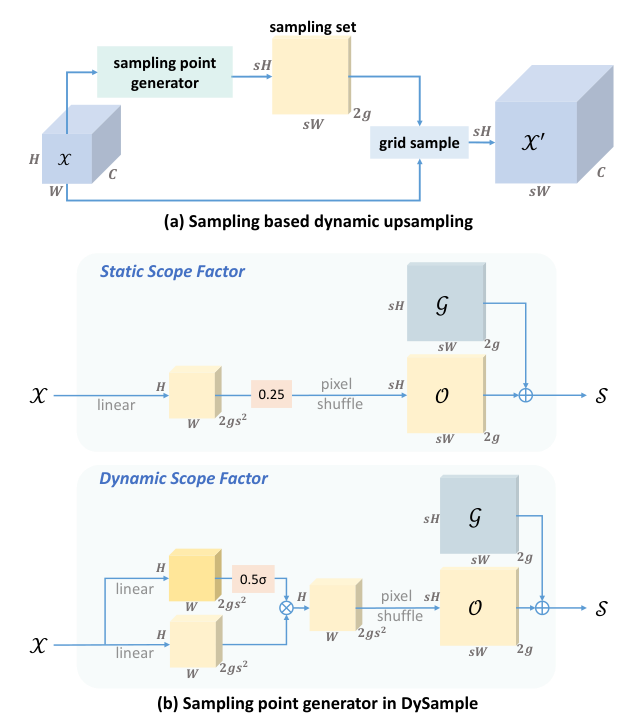}
  \caption{Sampling based dynamic upsampling and module designs in DySample}
  \label{fig:Dysample}
\end{figure}

The Figure \ref{fig:Dysample} illustrates the sampling-based dynamic upsampling approach and module design in Dysample. The diagram comprises two main components:

Sampling-Based Dynamic Upsampling (Fig. \ref{fig:Dysample}a): This section demonstrates the process of generating an upsampled feature map $(X^{\prime})$ from the input feature (X). Initially, a sampling set (S) is created by a sampling point generator. The grid\_sample function is then employed to resample the input feature (X) based on this set, producing the final upsampled output $(X^{\prime})$.

Sampling Point Generator in Dysample (Fig. \ref{fig:Dysample}b): This part elaborates on the two methodologies for generating sampling points: the Static Range Factor and the Dynamic Range Factor.

Static Range Factor: The offset (O) is generated by processing the input through a linear layer combined with a pixel shuffle operation, regulated by a fixed range factor. This offset (O) is subsequently added to the original grid positions (G) to form the sampling set (S).

Dynamic Range Factor: In addition to the linear layer and pixel shuffle, this method incorporates a dynamic range factor. A range factor is first generated and is then used to modulate the offset (O). Here, the $\sigma$ denotes the Sigmoid activation function, which is applied in generating this dynamic range factor.

\vspace{5pt}

\section{Experimental Results and Analysis}

\subsection{Experimental Environment and Hyperparameter Settings}
All experiments in this study were conducted on a Linux operating system, with detailed specifications provided in Table \ref{table:experimental_Env}. And, model training and validation were conducted based on the Ultralytics YOLO framework. The main hyperparameter settings used are shown in the Table \ref{tab:hyperparameters}.

\begin{table}
\centering
\caption{Experimental Environment}
\small
\begin{tabular}{ll}
\hline
\textbf{Item} & \textbf{Configuration} \\
\hline
CPU & Intel(R) Core(TM) i9-14900HX \\
GPU & NVIDIA GeForce RTX H800 \\
GPU Memory & 80G \\
Pytorch & Pytorch 2.4.1 \\
Python & Python 3.10.20 \\
CUDA & CUDA 11.8 \\
Utralytics & 8.3.170 \\
\hline
\label{table:experimental_Env}
\end{tabular}
\end{table}

\begin{table}[tb]
\centering
\caption{Training Hyperparameter Settings}
\label{tab:hyperparameters}
\begin{tabular}{p{0.3\linewidth}p{0.2\linewidth}p{0.4\linewidth}}
\hline
\textbf{Hyperparameter} & \textbf{Value} & \textbf{Description} \\
\hline
Epochs & 150 & Total number of training epochs \\
Batch Size & 128/96 & Number of images processed per batch in two runs \\
Initial Learning Rate & 0.01 & Starting learning rate for training \\
Final Learning Rate & 0.0001 & Final learning rate (lr0 × lrf) \\
Optimizer & SGD & Automatically select the optimal optimizer \\
Momentum & 0.937 & SGD momentum/Adam beta1 \\
Weight Decay & 0.0005 & Regularization parameter \\
Data Loading Workers & 64/48 & Number of parallel data loading threads in two runs\\
\hline
\end{tabular}
\end{table}

\subsection{Experimental Dataset}
This paper constructs FireRescue, the first large-scale object detection dataset designed for comprehensive firefighting and rescue scenarios, comprising 15980 images with 32000 high-quality annotations. The majority of the data is sourced from real fire rescue operations and training exercises worldwide, containing 8 categories of targets including 5 types of common fire rescue vehicles, firefighters, smoke, and flames.

\subsection{Ablation Study}
To validate the effectiveness of the individual improvements proposed in this paper for the enhanced fire rescue scene detection method, a comprehensive ablation study was conducted. All experiments were strictly controlled to ensure consistency in model parameters. The results of the ablation study are presented in Table \ref{table:ablation_simple}, where "+A" indicates the incorporation of the A2C2f-MCA improvement, "+B" denotes the addition of the A2C2f-MCEA improvement, and "+C" represents the integration of the "Dysample" module.

As evidenced in Table \ref{table:ablation_simple}, the effectiveness of each proposed improvement is validated. The introduction of the MCA module enhances cross-dimensional information interaction with a negligible increase in computational cost, resulting in gains of +1.26\% in mAP50, +1.78\% in mAP50-95, +7.83\% in Precision, +3.00\% in Recall,  thereby validating its contribution to detection performance. The proposed MCEA module, an enhancement to the MCA structure, delivers significant performance gains—+1.86\% mAP50, +1.84\% mAP50-95, +1.24\% Precision, and +2.94\% Recall—while requiring almost no increase in parameters or computational cost. Finally, replacing the conventional upsampler with the dynamic Dysample operator effectively mitigates class imbalance and directs greater attention to hard examples, which in turn enhances the detection accuracy for small and blurred objects. This final modification yields a further increase of +2.62\% in mAP50, +2.26\% in mAP50-95, +11.98\% in Precision, +3.12\% in Recall, confirming its substantial benefit to overall model performance.

\begin{table*}[htb]
\centering
\caption{Ablation study on the FireRescue dataset}
\label{table:ablation_simple}
\setlength{\tabcolsep}{12pt} % 增加列间距
\begin{tabular}{lccccccc}
\hline
\multicolumn{1}{c}{\textbf{Model}} & \textbf{mAP50} & \textbf{mAP50-95} & \textbf{Precision} & \textbf{Recall} & \textbf{Params(M)} & \textbf{GFLOPs}\\ 
\hline
baseling    & 0.65869 & 0.38103 & 0.79117 & 0.62561 & 2.6 & 6.5 \\
+A          & 0.67132 & 0.39883 & 0.86948 & 0.65565 & 2.5 & 6.2 \\ 
+B          & 0.67732 & 0.39946 & 0.80366 & 0.65504 & 2.5 & 6.2 \\
+C          & 0.67680 & 0.40130 & 0.79603 & 0.64308 & 2.6 & 6.5 \\
+A+C        & 0.68388 & 0.40349 & 0.84620 & 0.65169 & 2.5 & 6.2 \\ 
+B+C        & 0.68493 & 0.40365 & 0.91106 & 0.65684 & 2.5 & 6.2 \\
\hline
\end{tabular}
\end{table*}

\subsection{Comparative Experiments}

\begin{table*}[htb]
\centering
\caption{Performance Comparison of Different Models on the FireRescue Dataset}
\label{table:perf-comparison}
\setlength{\tabcolsep}{12pt} % 增加列间距
\begin{tabular}{lcccccc}
\hline
\multicolumn{1}{c}{\textbf{Model}} & \textbf{mAP50} & \textbf{mAP50-95} & \textbf{Precision} & \textbf{Recall} & \textbf{Params(M)} & \textbf{GFLOPs}\\ 
\hline
YOLOv3-tiny & 0.61258 & 0.34372 & 0.7831 & 0.58579 & 2.5 & 7.1  \\
YOLOv5nu & 0.64649 & 0.37861 & 0.7688 & 0.60921 & 2.6 & 7.7   \\
YOLOv6n & 0.63701  & 0.37049 & 0.79729 & 0.5851 & 4.2 & 11.7 \\
YOLOv8n & 0.66694 & 0.39639 & 0.79785 & 0.63559 & 3.5 & 10.5 \\
YOLOv9t & 0.54452 & 0.31304 & 0.76716 & 0.54023 & 2.0 & 7.7  \\
YOLOv10n & 0.66728  & 0.3971 & 0.8188 & 0.63906 & 2.3 & 6.7  \\
YOLOv11n & 0.53933  & 0.31657 & 0.77228 & 0.53967 & 2.6 & 6.4  \\
YOLOv12n & 0.65869  & 0.38103 & 0.79117 & 0.62561 & 2.6 & 6.5  \\
Mamba-YOLOt & 0.66546 & 0.39783 & 0.78829 & 0.62745 & 6.0 & 13.6  \\
FRS-YOLOn(ours)  & \textbf{0.68493} & \textbf{0.40365} & \textbf{0.81106} & \textbf{0.65684} & 2.5 & 6.2  \\
\hline
RTDetr & 0.55143 & 0.31893 & 0.7459 & 0.56918 & 320.0 & 103.5   \\
YOLOv3 & 0.67386  & 0.4077 & 0.80284 & 0.64746 & 103.6 & 282.2  \\
YOLOv5su & 0.68456 & 0.40947 & 0.78673 & 0.66119 & 9.1 & 24.1   \\
YOLOv6s & 0.65418 & 0.38075 & 0.79444 & 0.61829 & 16.3 & 43.9 \\
YOLOv8s & 0.68854  & 0.36942 & 0.79609 & 0.64973 & 11.4 & 29.7 \\
YOLOv9s & 0.69309 & 0.41547 & 0.7823 & 0.665 &  7.2  &  26.7  \\
YOLOv10s & 0.68024 & 0.40995 & 0.78448 &  0.6569 & 8.0 & 24.8 \\
YOLOv11s & 0.68558 & 0.40973 & 0.81349 & 0.66232 & 9.4 & 21.5  \\
YOLOv12s & 0.69381 & 0.41192 & 0.7852 & 0.6741 & 9.2 & 21.2 \\
Mamba-YOLOb & 0.68423 & 0.40529 & \textbf{0.81729} & 0.64268 & 21.8 & 49.6 \\
FRS-YOLOs & \textbf{0.70514} & \textbf{0.41199} & 0.81457 & \textbf{0.67811} & 9.1 & 20.9 \\
\hline
\end{tabular}
\end{table*}

In this study, we enhance YOLOv12 by integrating novel designed modules. To comprehensively evaluate the effectiveness of our proposed method, we conduct two sets of comparative experiments based on model scale. First, the proposed module is incorporated into two different scales of YOLOv12, namely YOLOv12n and YOLOv12s, resulting in two new architectures: FRS-YOLOn and FRS-YOLOs. Subsequently, a rigorous performance comparison is carried out. The decision to employ both nano (n) and small (s) scale models for evaluation is twofold. First, it aims to demonstrate the generalizability and scalability of our proposed module. A common pitfall in architectural innovation is that improvements may be effective only at a specific model capacity. By validating our module across two distinct scales, we rigorously show that its benefits are not merely incidental to a particular parameter count but are a consistent and transferable enhancement. Second, this approach provides a more comprehensive performance characterization. The nano-scale models, with their extreme efficiency, are critical for applications in resource-constrained environments. Improvements here are measured by the gain in performance per parameter. Conversely, the small-scale models offer a higher baseline performance, allowing us to evaluate whether our module can provide a significant boost even when the base architecture is already more powerful. This dual-level assessment offers complete evidence of our module's practical utility across different performance-efficiency trade-offs.

The detailed performance comparison results of different models are shown in Table \ref{table:perf-comparison}, from which the following conclusions can be drawn:

The benefits of our module manifest differently across model scales, highlighting its versatile utility. For the computation-bound nano-scale models, the primary advantage lies in a remarkable +2.62\% increase in mAP50, +2.26\% increase in mAP50-95, +11.98\% increase in Precision, +3.12\% increase in Recall, crucial for reducing missed detections in resource-constrained deployments. Conversely, for the more powerful small-scale models, our module pushes the performance ceiling further, achieving a significant +1.13\% boost in mAP50, +2.93\% boost in Precision, +0.4\% boost in Recall, which is vital for high-accuracy applications.

Crucially, our enhanced model FRS-YOLOn outperforms the original YOLOv12n by a clear margin, while both models possess an identical number of parameters. This provides direct and fair evidence that the improvement stems from our architectural innovation, not simply an increase in model size. The same trend holds for the small-scale group, where FRS-YOLOs consistently surpasses YOLOv12s.

Through comprehensive analysis, our final model FRS-YOLO achieves the optimal balance in overall performance among all compared models.

\subsection{Visualization of Experimental Results}
To provide an intuitive comparison of the detection performance between the proposed improved model and the baseline model, challenging images were selected from the test set for comparative evaluation. The superiority of the improved model is demonstrated through two visualization methods: detection results and detection heatmaps.

\subsubsection{Visualization of Detection Results}
This subsection qualitatively demonstrates the performance improvement of our proposed FRS-YOLO across three challenging scenarios through comparative experiments. For a comprehensive evaluation, we compare the "n" scale models in the first and third experimental groups to assess their performance under lightweight configurations, while the "s" scale models are used in the second group to investigate their detection capability with higher computational budgets.

\begin{figure*}[!t]
  \centering
  % 第一张图
  \begin{minipage}[b]{0.32\linewidth}
    \centering
    \includegraphics[width=\textwidth]{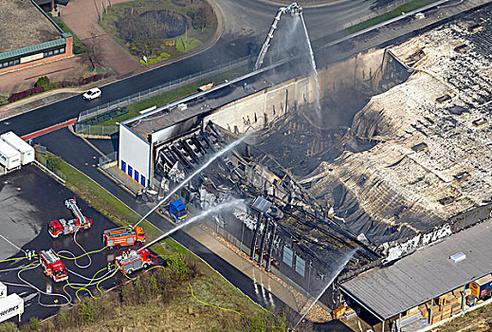}
    \subcaption{Original Image}
    \label{fig:detection2-original}
  \end{minipage}
  \hfill
  % 第二张图
  \begin{minipage}[b]{0.32\linewidth}
    \centering
    \includegraphics[width=\textwidth]{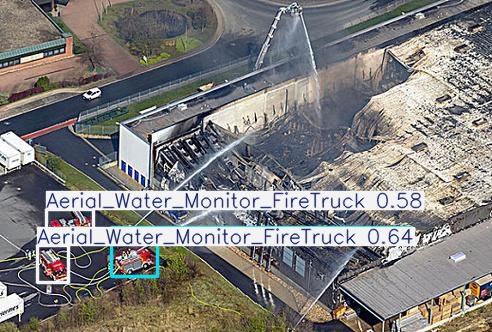}
    \subcaption{YOLOv12n}
    \label{fig:detection2-yolov12}
  \end{minipage}
  \hfill
  % 第三张图
  \begin{minipage}[b]{0.32\linewidth}
    \centering
    \includegraphics[width=\textwidth]{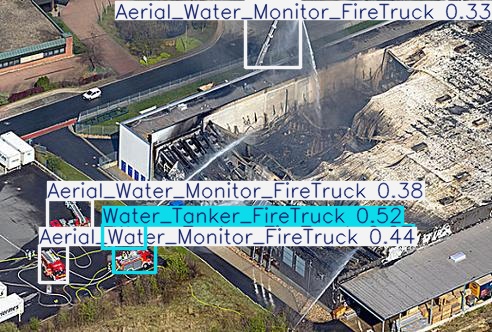}
    \subcaption{FRS-YOLOn}
    \label{fig:detection2-FRS-YOLO}
  \end{minipage}
  
  \caption{Comparative Analysis of Model Performance in Small and Occluded Object Detection Capability}
  \label{fig:detection-comparison-one}
\end{figure*}

\begin{figure*}[!t]
  \centering
  % 第一张图
  \begin{minipage}[b]{0.32\linewidth}
    \centering
    \includegraphics[width=\textwidth]{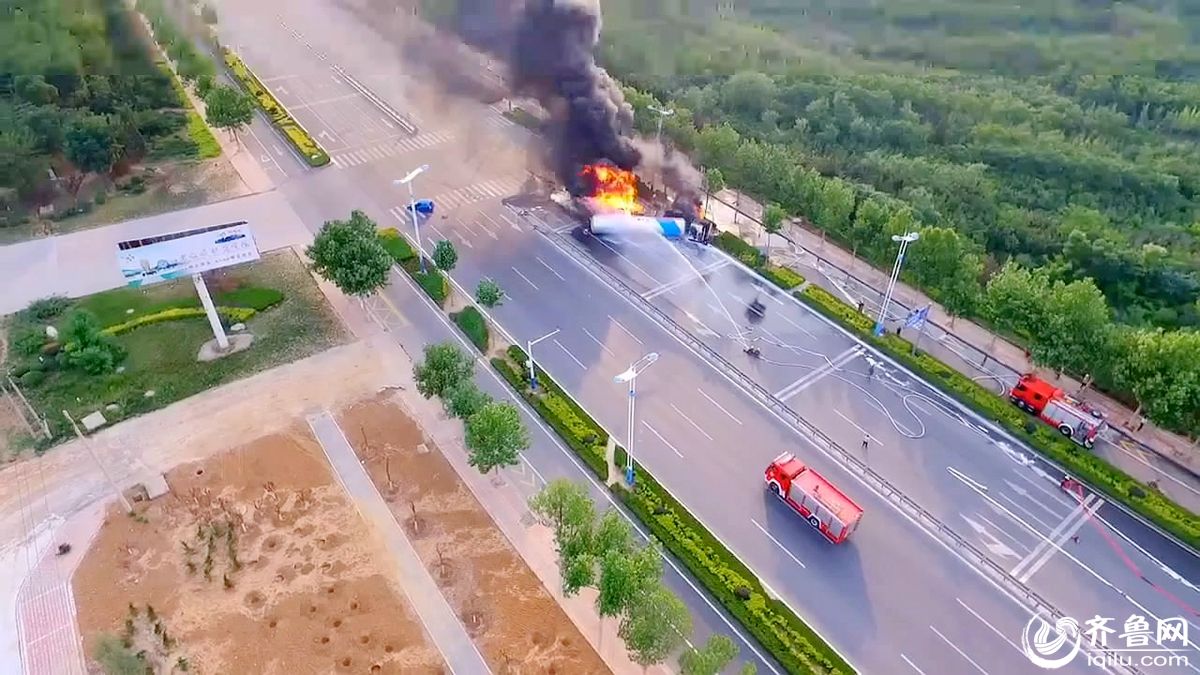}
    \subcaption{Original Image}
    \label{fig:detection2-original}  % 保持原始标签
  \end{minipage}
  \hfill
  % 第二张图
  \begin{minipage}[b]{0.32\linewidth}
    \centering
    \includegraphics[width=\textwidth]{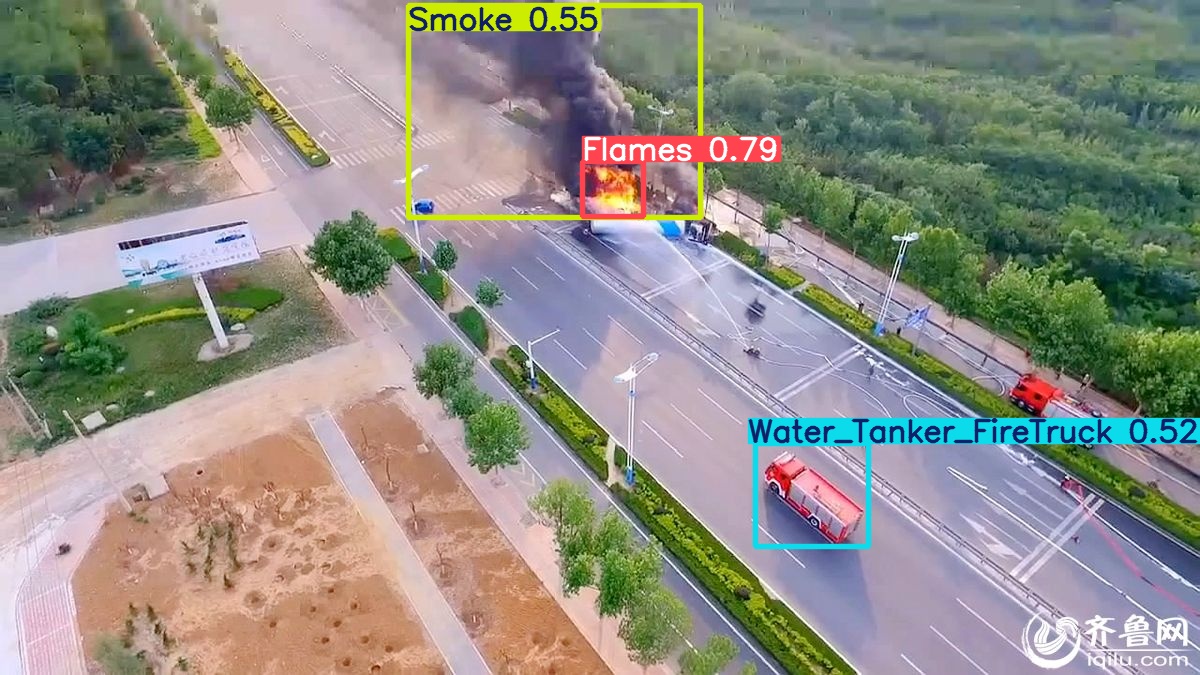}
    \subcaption{YOLOv12n}
    \label{fig:detection2-yolov12}   % 保持原始标签
  \end{minipage}
  \hfill
  % 第三张图
  \begin{minipage}[b]{0.32\linewidth}
    \centering
    \includegraphics[width=\textwidth]{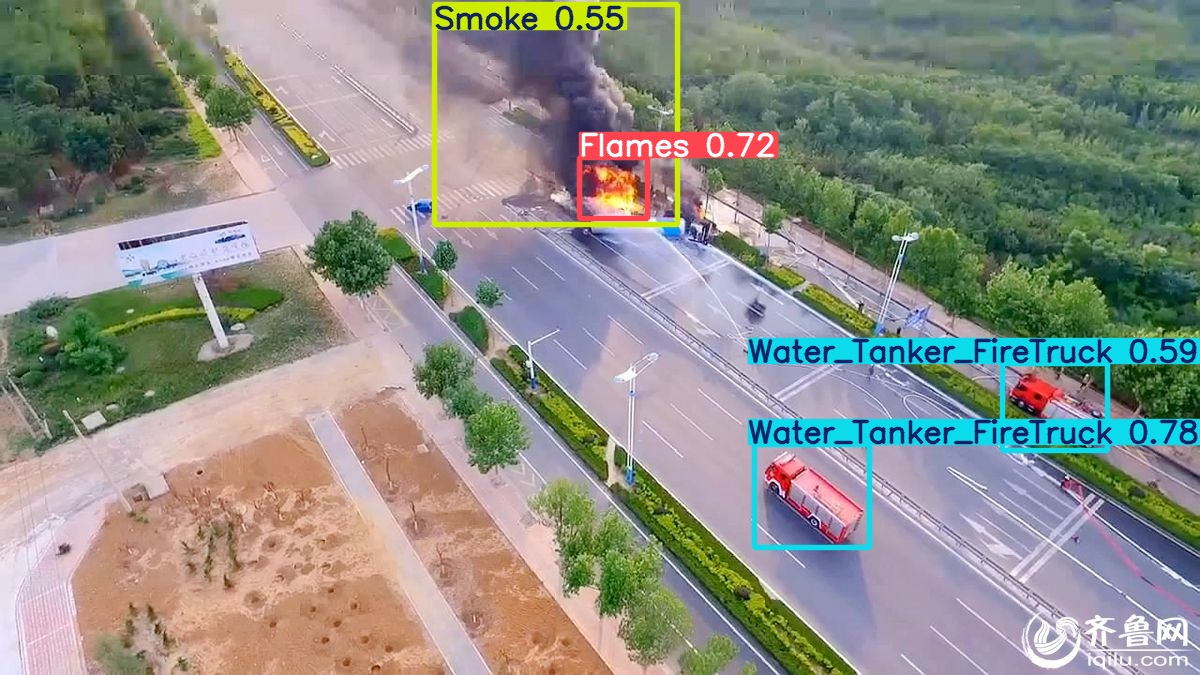}
    \subcaption{FRS-YOLOn}
    \label{fig:detection2-FRS-YOLO}  % 保持原始标签
  \end{minipage}
  
  \caption{Comparative Analysis of Model Performance in Detection Robustness under Extreme Visual Interference}
  \label{fig:detection-comparison-two}
\end{figure*}

\begin{figure*}[!t]
  \centering
  % 第一张图
  \begin{minipage}[b]{0.32\linewidth}
    \centering
    \includegraphics[width=\textwidth]{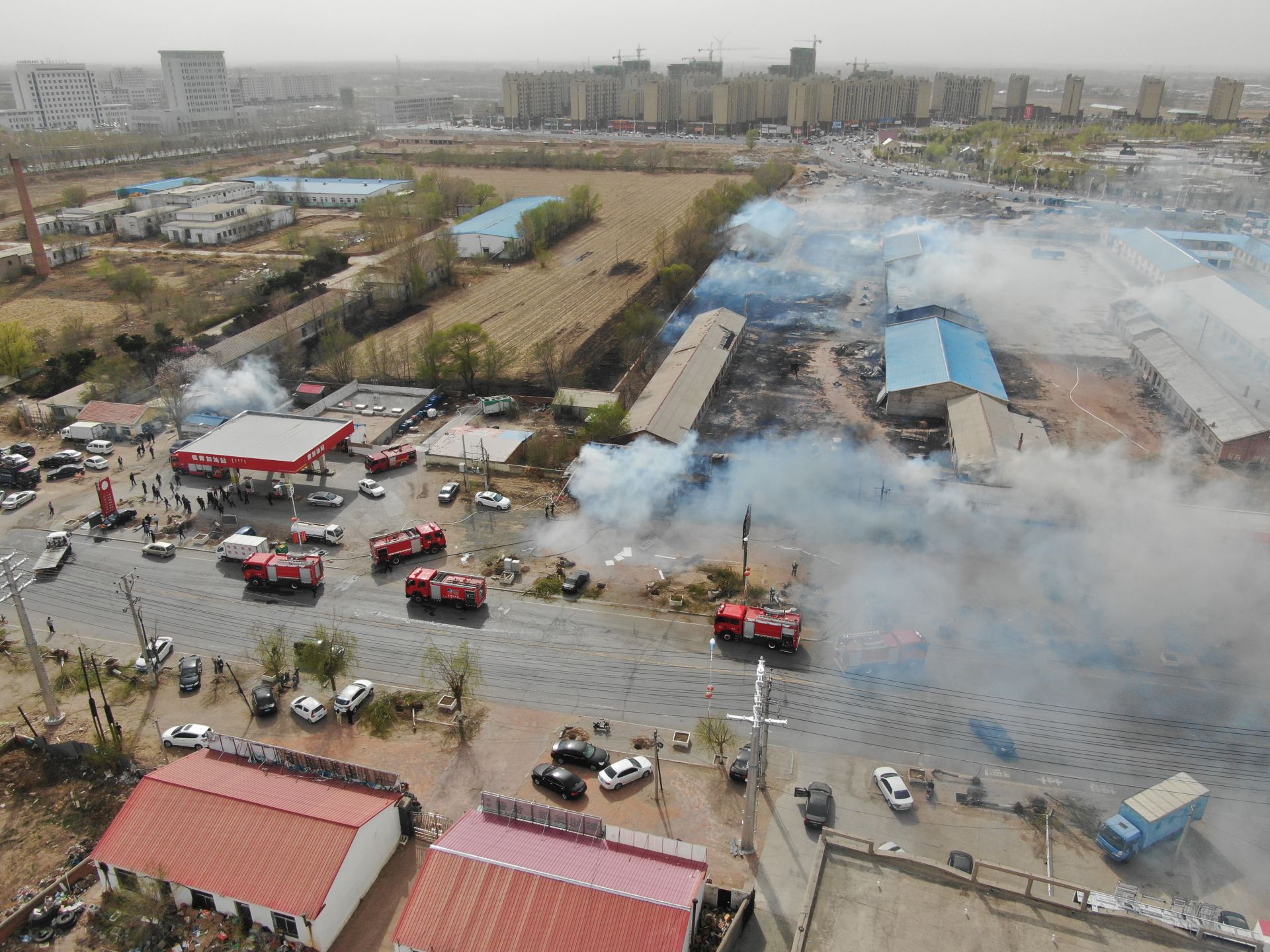}
    \subcaption{Original Image}
    \label{fig:detection2-original}
  \end{minipage}
  \hfill
  % 第二张图
  \begin{minipage}[b]{0.32\linewidth}
    \centering
    \includegraphics[width=\textwidth]{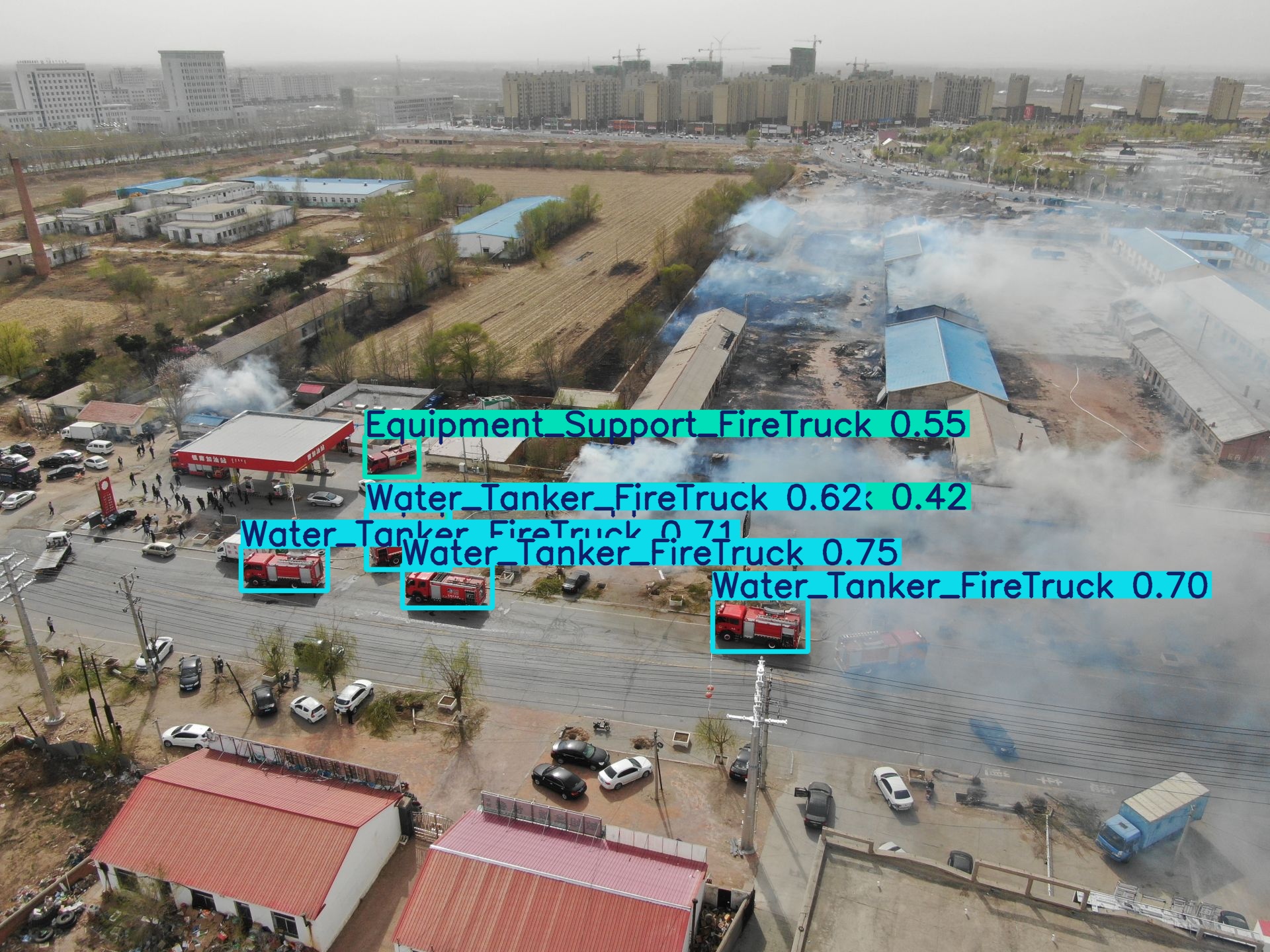}
    \subcaption{YOLOv12n}
    \label{fig:detection2-yolov12}
  \end{minipage}
  \hfill
  % 第三张图
  \begin{minipage}[b]{0.32\linewidth}
    \centering
    \includegraphics[width=\textwidth]{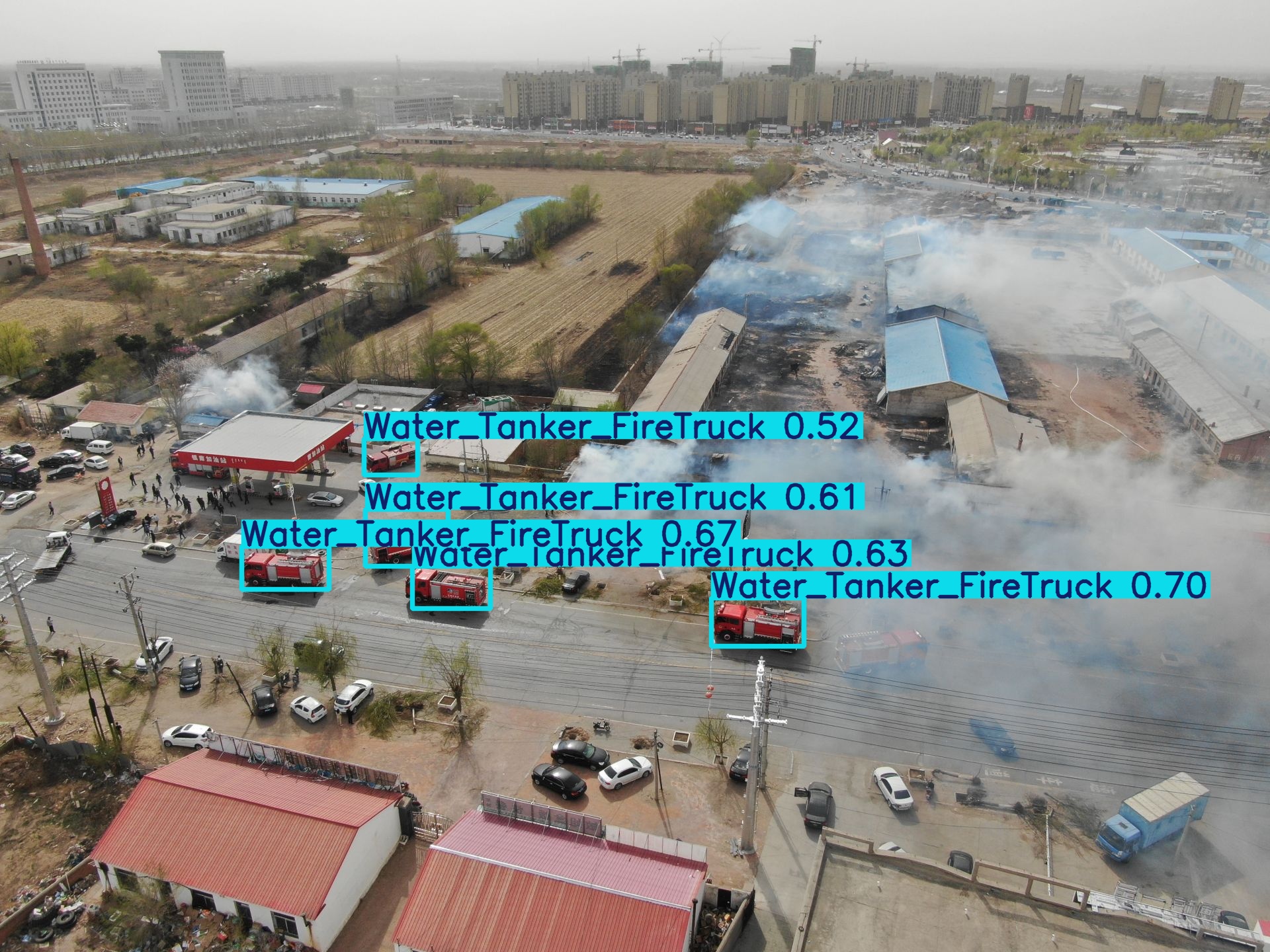}
    \subcaption{FRS-YOLOn}
    \label{fig:detection2-FRS-YOLO}
  \end{minipage}
  
  \caption{Comparative Analysis of Model Performance in Multi-Target Discrimination Capability in Dense Scenes}
  \label{fig:detection-comparison-three}
\end{figure*}

(1) Small and Occluded Object Detection Capability

As shown in Figure \ref{fig:detection-comparison-one}, the first scenario focuses on detecting small and partially occluded objects. In a real fire disposal scene with complex backgrounds, the environment contains fire trucks that appear extremely small in pixel area due to long distances, along with vehicles partially occluded by other objects. The baseline YOLOv12n model exhibits significant missed detections in this environment. In contrast, our FRS-YOLOn model successfully detects all targets. Crucially, the model can accurately infer and detect severely occluded fire trucks through local distinctive features such as the folding arms. This demonstrates the powerful capability of our MCEA module in capturing and reasoning about subtle features and critical local cues.

(2) Detection Robustness under Extreme Visual Interference

The second comparative experiment, shown in Figure \ref{fig:detection-comparison-two}, aims to evaluate model performance under extreme visual interference. The scenario involves a vehicle fire on a highway, with strong interfering factors such as dense smoke and flames that cause image blurring and contrast reduction. The baseline YOLOv12s model shows substantial missed detections for critical targets like fire trucks in this environment, indicating insufficient anti-interference capability. Our FRS-YOLOs model demonstrates stronger robustness, with significantly improved recall rates for key targets. It should be noted that our model shows slightly lower precision in flame detection compared to the baseline. We attribute this to an effective trade-off where the model sacrifices some flame detection precision to achieve substantial improvement in recall rates for core rescue targets (e.g., fire trucks) under extreme interference, which is of great practical significance in emergency rescue operations.

(3) Multi-Target Discrimination Capability in Dense Scenes

The third experiment, presented in Figure \ref{fig:detection-comparison-three}, evaluates performance in scenarios with multiple coexisting vehicles. The baseline YOLOv12n model produces false detections in this setting, misclassifying non-emergency vehicles as fire trucks. Our FRS-YOLOn model demonstrates superior discrimination capability, accurately identifying all targets. This indicates that the improved model possesses enhanced feature discriminability for visually similar vehicle categories, effectively reducing false positive rates in dense scenes.

\subsubsection{Heatmap Visualization and Analysis}

\begin{figure*}[!t]
  \centering
  
  % 使用tabular进行更精确的3列控制
  \begin{tabular}{c c c}
    % ========== 第一列：Original Images ==========
    \begin{minipage}[b]{0.32\linewidth}
      \centering
      \textbf{Original Images}\\[2mm]
      
      % 第一行图片
      \includegraphics[width=\linewidth, height=4.5cm, keepaspectratio]{./figs/contrast-map/5/5.jpg}\\[2mm]
      
      % 第二行图片
      \includegraphics[width=\linewidth, height=4.5cm, keepaspectratio]{./figs/contrast-map/1/1.jpeg}\\[2mm]

      % 第三行图片
      \includegraphics[width=\linewidth, height=4.5cm, keepaspectratio]{./figs/contrast-map/9/9.jpeg}\\[2mm]

    \end{minipage}
    &
    % ========== 第二列：YOLOv12 ==========
    \begin{minipage}[b]{0.32\linewidth}
      \centering
      \textbf{YOLOv12}\\[2mm]
      
      % 第一行图片
      \includegraphics[width=\linewidth, height=4.5cm, keepaspectratio]{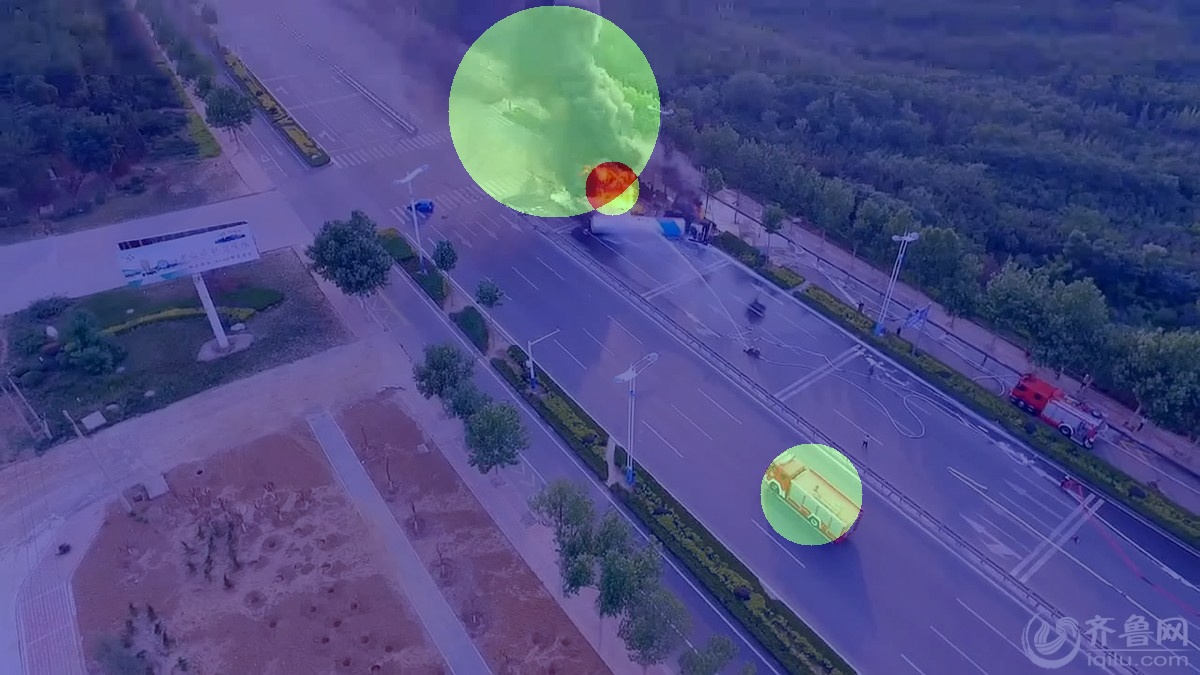}\\[2mm]
        
      % 第二行图片
      \includegraphics[width=\linewidth, height=4.5cm, keepaspectratio]{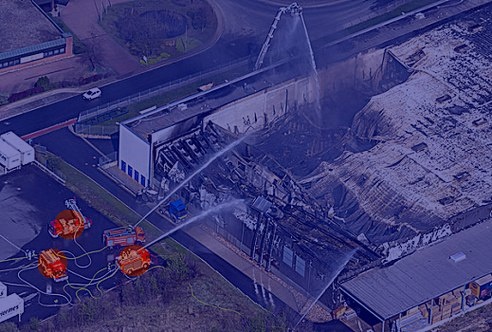}\\[2mm]

      % 第三行图片
      \includegraphics[width=\linewidth, height=4.5cm, keepaspectratio]{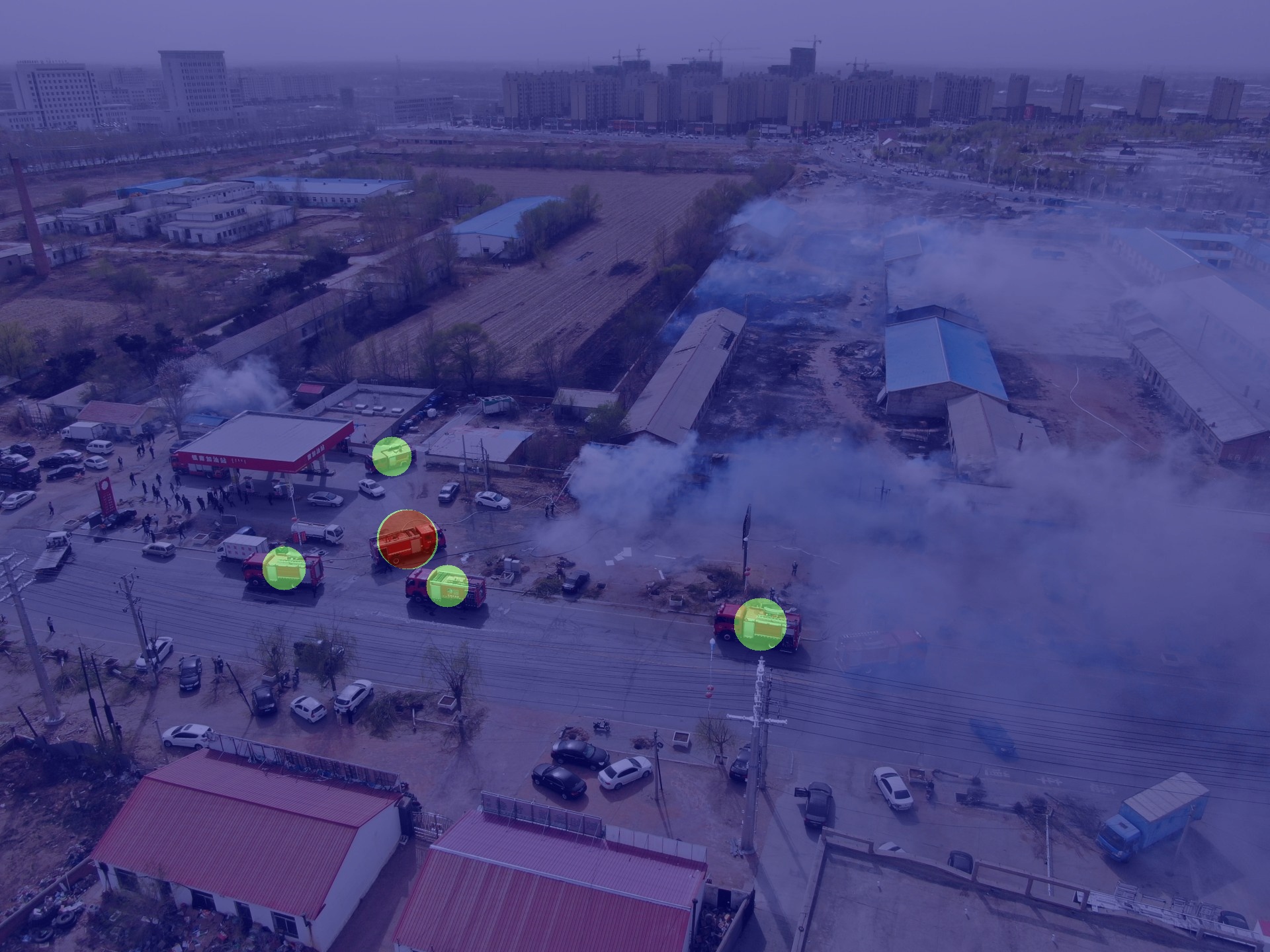}\\[2mm]

    \end{minipage}
    &
    % ========== 第三列：FRS-YOLO ==========
    \begin{minipage}[b]{0.32\linewidth}
      \centering
      \textbf{FRS-YOLO}\\[2mm]
      
      % 第一行图片
      \includegraphics[width=\linewidth, height=4.5cm, keepaspectratio]{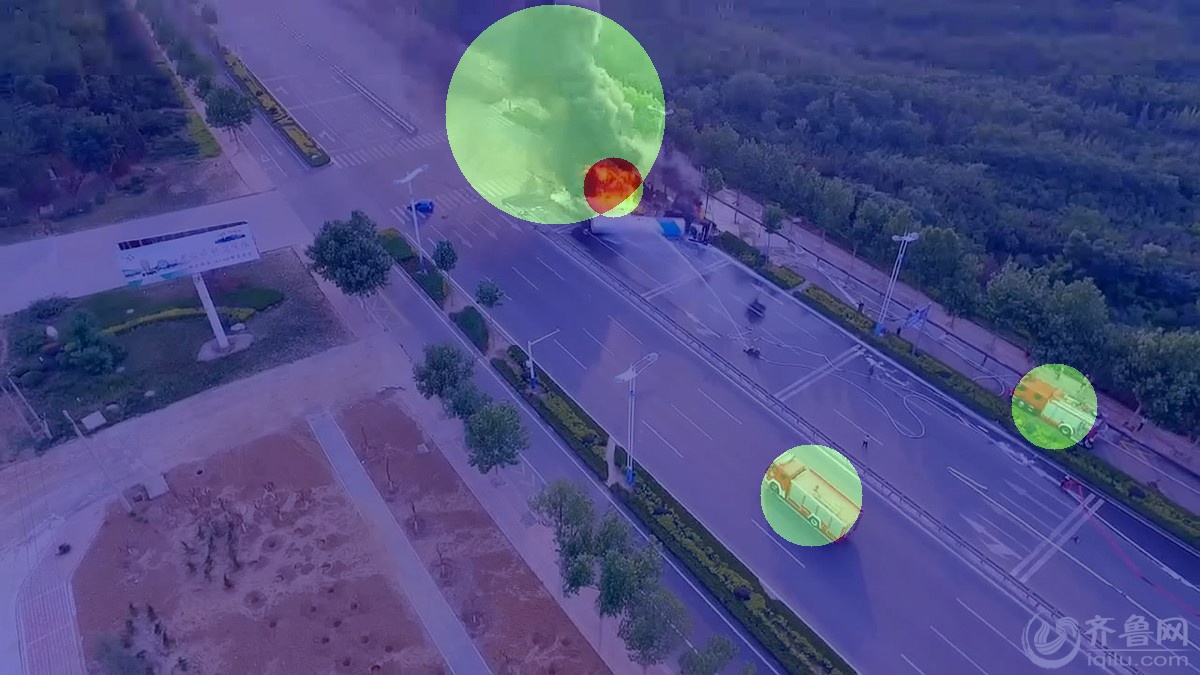}\\[2mm]

      % 第二行图片
      \includegraphics[width=\linewidth, height=4.5cm, keepaspectratio]{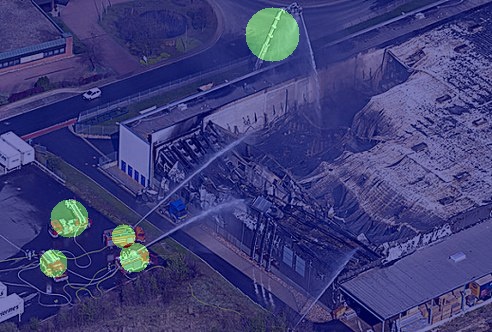}\\[2mm]
      
      % 第三行图片
      \includegraphics[width=\linewidth, height=4.5cm, keepaspectratio]{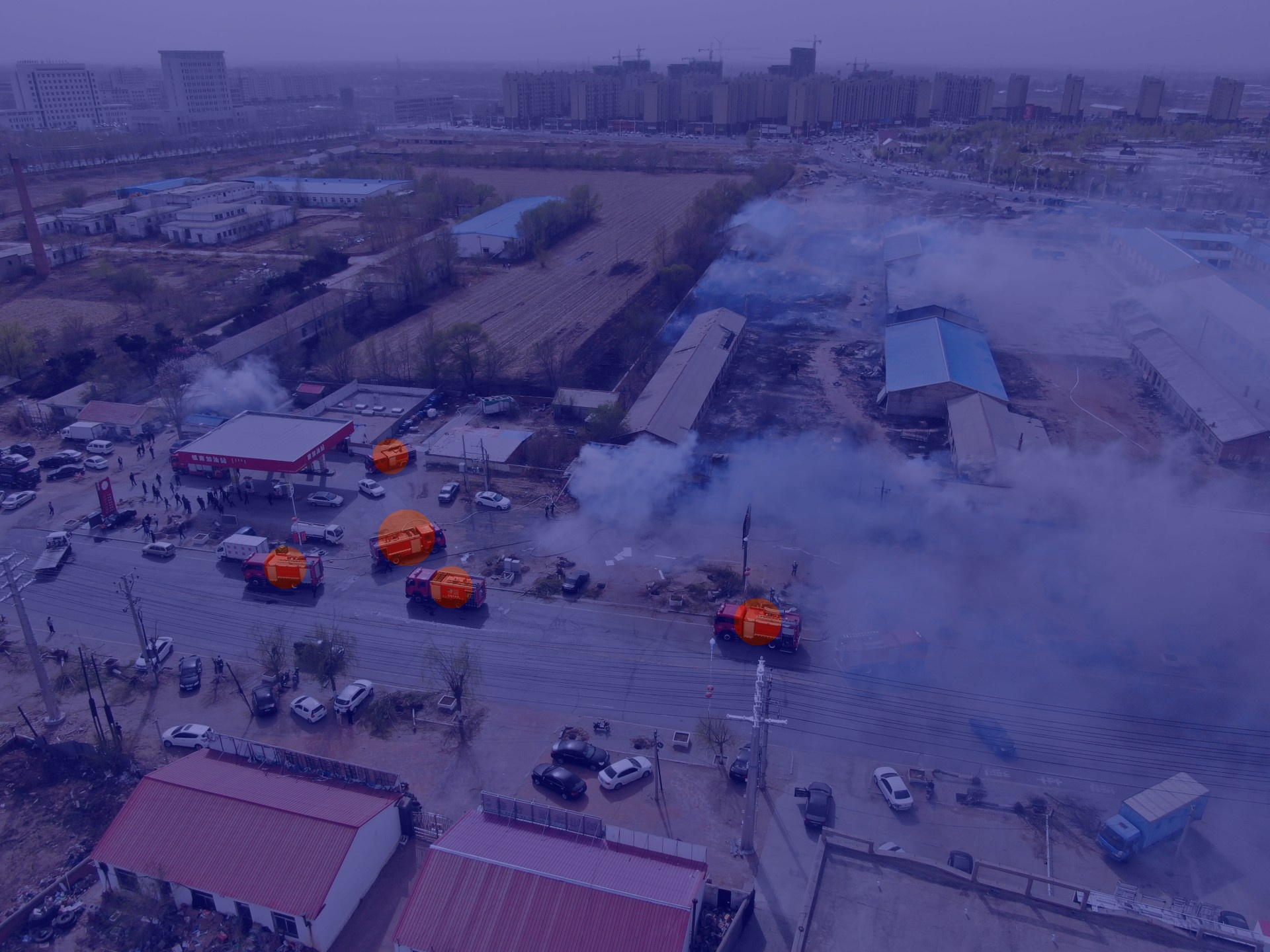}\\[2mm]

    \end{minipage}
  \end{tabular}
  
  \vspace{4mm}
  \caption{Comparison of detection results between YOLOv12n and FRS-YOLO}
  \label{fig:contrast-heatmaps}
\end{figure*}

To gain deeper insight into the models' internal decision-making processes and to explain the performance differences observed in the detection results (Section 4.5.1), we employ Grad-CAM to generate corresponding heatmaps for the same three sets of scenarios, visualizing the models' attention regions during inference. Figure \ref{fig:contrast-heatmaps} presents the heatmap visualizations of both the baseline model and the improved model when processing challenging images from the test set. In the heatmaps, deeper red coloration indicates higher model attention to the corresponding feature regions.

(1) Analysis of Attention Regions and Missed Detections

As shown in Figure\ref{fig:contrast-heatmaps}, the heatmaps for our model in the first and second scenarios show that its attention is highly concentrated on all fire trucks, including key components of occluded vehicles (e.g., the folding arms), with strong and focused activation responses. In contrast, the heatmaps of the baseline YOLOv12n clearly reveal the reason for its missed detections: the model produces almost no significant activation response in the regions of the missed fire trucks, as its attention is erroneously dispersed to the background or meaningless areas. This fundamentally confirms that our MCEA module more effectively guides the model to focus on crucial targets in the image rather than irrelevant background.

(2) Visual Evidence of Category Confusion

The heatmaps from the third experimental group as shown in Figure \ref{fig:contrast-heatmaps} provide direct evidence for the false detections produced by the baseline model. It can be observed that on the falsely detected non-emergency vehicles, the baseline model's heatmap displays distinct areas of color overlap (e.g., a mixture of red and green). This indicates that different internal channels (likely corresponding to different categories) within the model generated high responses in these areas, leading to ambiguity in the model's category judgment for the target and ultimately resulting in false detection. Conversely, our model's heatmap shows clearly distinguished activation areas and pure colors for different categories of targets, demonstrating that the improved model possesses stronger feature discrimination capability and more definitive decision-making basis.

The heatmap analysis indicates that the performance improvement of our model stems from its more rational visual attention mechanism—enabling precise focus on critical targets while maintaining clear category discrimination. This provides an intrinsic explanation for the quantitative results in Section 4.5.1.

\section{Conclusion}
This paper tackles the challenges of object detection in complex fire rescue scenarios by introducing FRS-YOLO, an enhanced model built upon YOLOv12. A key contribution of this work is the creation of the first comprehensive dataset specifically designed for fire rescue scenarios, providing a valuable benchmark for future research in this domain. Additional contributions include the design of a Multi-dimensional Collaborative Enhancement Attention (MCEA) module, which incorporates global average pooling, standard deviation pooling, and max pooling to enrich feature representation, along with the reconstruction of the A2C2f module in the backbone network to strengthen the focus on discriminative regions. Furthermore, we introduce a dynamic sampler (Dysampler) that adaptively enhances the model's attention on hard examples, significantly improving detection accuracy for small and blurry targets which are particularly challenging in fire rescue environments. Experimental evaluations demonstrate that the proposed approach achieves substantial improvements in precision, mAP50, mAP50-90 and recall over the baseline model, while preserving desirable lightweight properties. 

\section{Acknowledgement}
\label{sec:Acknowledgement}
This work was supported in part by the National Natural Science Foundation of China under Grant (62271119, 62071086), the Key Research and Development Project of Hainan Province under Grant ZDYF2024(LALH)003, and the Natural Science Foundation of Sichuan Province under Grant (2023NSFSC1972, 2025ZNSFSC0475).

\bibliography{mybibfile}%%bib

% \end{thebibliography}
\end{document}